\documentclass{llncs}
\usepackage{llncsdoc}
\usepackage{stmaryrd}
\usepackage{amsmath}
\usepackage{amssymb}
\usepackage{graphicx,times} 
\usepackage{bm}
\usepackage[tight,footnotesize]{subfigure}
\usepackage[dvips]{epsfig}
\usepackage[tight,footnotesize]{subfigure}
\usepackage{multirow}
\usepackage{algpseudocode}
\usepackage{float}
\usepackage{algorithm}
\usepackage[mathlines]{lineno}

\begin{document}

\title{Locally linear representation for image clustering}

\author{Liangli Zhen, Zhang Yi, Xi Peng and Dezhong Peng}
\institute{Machine Intelligence Laboratory, College of Computer Science, Sichuan University, \\Chengdu 610065, P.~R.~China\\\email{llzhen@outlook.com, \{zhangyi,~pengdz\}@scu.edu.cn,\\pangsaai @gmail.com}}
\maketitle

\abstract{The construction of similarity graph plays an essential role in the spectral clustering algorithm. There exist two popular schemes to construct a similarity graph, i.e., the pairwise distance-based scheme and the linear representation-based scheme. It is notable that the above schemes  suffered from some limitations and drawbacks, respectively. Specifically, the pairwise distance-based scheme is sensitive to noises and outliers, while the linear representation-based scheme may incorrectly select inter-subspaces points to represent the objective point. These drawbacks  degrade the performance of the spectral clustering algorithms  greatly. To overcome these problems, the present letter proposes a novel scheme to construct the similarity graph, where the similarity computation among different data
points depends on both their pairwise distances and the linear representation relationships. This proposed scheme, called Locally Linear Representation (LLR), encodes each data point using a collection of data points that not only produce the minimal reconstruction error but also are close to the objective point, which makes it robust to noises and outliers, and avoid selecting inter-subspaces points to represent the objective point to a large extent.}

\section{Introduction}
\label{sec1} Spectral clustering is one of the most popular clustering algorithms, whose key is to build a similarity graph to describe the similarities among different data points~\cite{Cheng2010}. In the graph, each vertex denotes a data point, and the edge weight between two vertexes represents the similarity of the corresponding data points. Currently, there are two schemes to
calculate the similarity among data points, i.e., Pairwise Distance-based Scheme (PDS) and Linear Representation-based Scheme (LRS).  PDS computes the similarity between two points according
to the distance between two points, e.g., Laplacian Eigenmaps (LE)~\cite{Belkin2003}. On the other hand, LRS assumes that each data point could be denoted as a linear combination of some intra-subspace
points~\cite{Elham2013}. Based on this observation, this scheme uses the linear representation coefficients as a measure  of similarity. Recently, LRS has attracted more interests from the field of image clustering, since it capture the real structure of the data set better. Numerous clustering algorithms are developed based on LRS, such as Locally Linear Embedding (LLE)~\cite{Roweis2000}, Sparse Subspace Clustering (SSC)~\cite{Elham2013} and Low Rank Representation (LRR)~\cite{Liu2013}.

It is notable that the above-mentioned similarity computation schemes suffer respectively  from some limitations. Specifically, Pairwise Distance-based Scheme (PDS) is sensitive to noises and
outliers, because it only depends on the distance between the two considered data points, and ignores the global structure of the whole data set. Fig.~\ref{pairwise} illustrates the disadvantages
of PDS. On the other hand, Linear Representation-based Scheme (LRS) has the possibility that a data point is represented as a linear combination of the inter-subspace data points. Fig.~\ref{linear}  shows the drawbacks of  LRS. SSC~\cite{Elham2013} and LRR~\cite{Liu2013} overcome this problem to some extent by bringing a sparsity  constraint and a low-rank constraint into linear representation, but both of them are iterative algorithms with high computational complexity.

In order to overcome the above-mentioned problems, this letter presents a novel scheme to construct the similarity graph, where the similarity computation among different data points depends on not only their pairwise distances but also mutually linear representation relationships. The proposed scheme, called Locally Linear Representation (LLR), encodes each data point using a set of data points which produce the minimal  error, and are close to the objective point. Our developed scheme is more robust to noises and outliers than PDS. At the same time, being compared with LRS, it can effectively avoid selecting inter-subspaces points to represent the objective point. Moreover, the new scheme uses  an analytic solution to construct the similarity graph, and has lower computational complexity than the iterative methods, such as SSC and LRR.

\section{Locally Linear Representation}
\label{Sec2}

Our basic idea was derived from a theoretical result in manifold learning that a topological manifold is a topological space which is locally homeomorphic to an Euclidean space~\cite{Roweis2000}. It implies that in a subspace, mutually adjacent points can provide  the  linearly representation for  each other. This inspire us to construct the similarity graph by solving the following optimization problem:

For each point $\mathbf{x}_i, i=1, 2, \dots, n$,
\begin{equation}
\label{equ:optimal}
\begin{split}
\min_{\mathbf{c}_i} \hspace{1mm} {\lambda \|\mathbf{S}_i\mathbf{c}_i\|_2^2+(1-\lambda)\|\mathbf{x}_i-\mathbf{D}_i\mathbf{c}_i\|_2^2}\hspace{3mm}
\mathrm{s.t.} \hspace{1mm} \mathbf{1}^T\mathbf{c}_i=1,
\end{split}
\end{equation}
where $\mathbf{D}_i=[\mathbf{x}_1, \mathbf{x}_2, \dots, \mathbf{x}_{i-1}, \mathbf{0}, \mathbf{x}_{i+1}, \dots, \mathbf{x}_n]$ is a dictionary for $\mathbf{x}_i$, $\mathbf{S}_i$ is a diagonal matrix whose $j$-th diagonal element is the pairwise distance between $\mathbf{x}_i$ and the $j$-th data point in $\mathbf{D}_i$, $\mathbf{1}\in \mathbb{R}^m$ is a vector consists of ones, $\lambda \in [0, 1)$ is a balance parameter, and $\mathbf{c}_i \in \mathbb{R}^n$ is the representation coefficient of $\mathbf{x}_i$.

\begin{figure}[t]
\centering \subfigure[PDS]{ \label{pairwise}
\includegraphics[width=28mm]{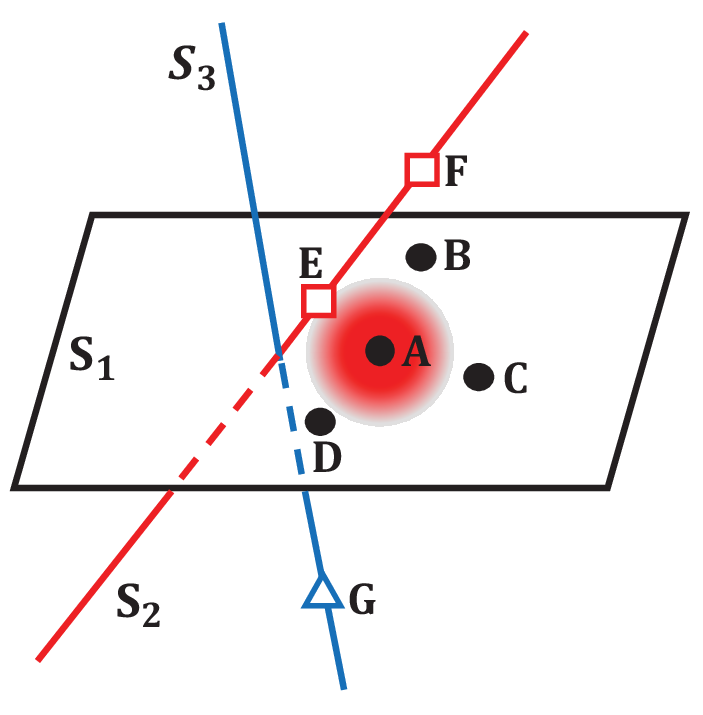}}
\subfigure[LRS]{ \label{linear}
\includegraphics[width=28mm]{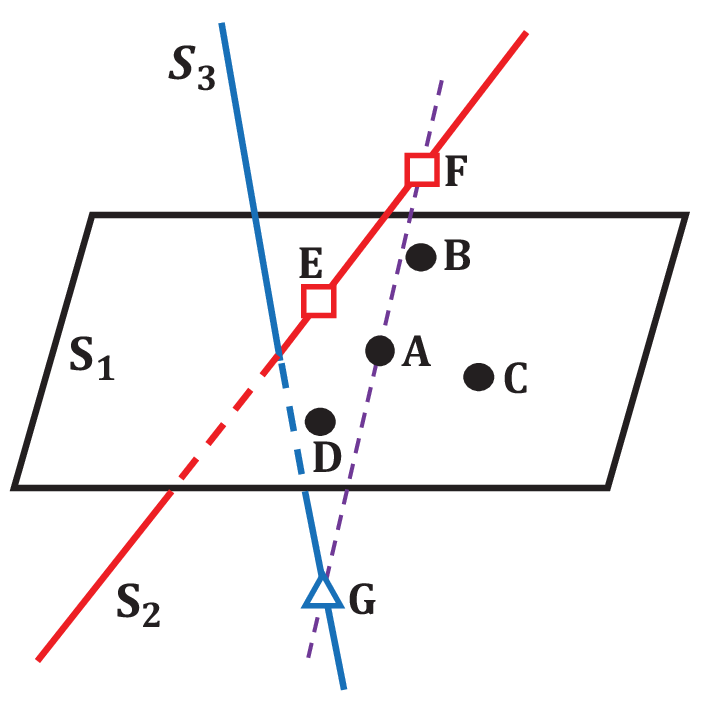}}
\subfigure[Our method]{
\label{locallylinear}
\includegraphics[width=28mm]{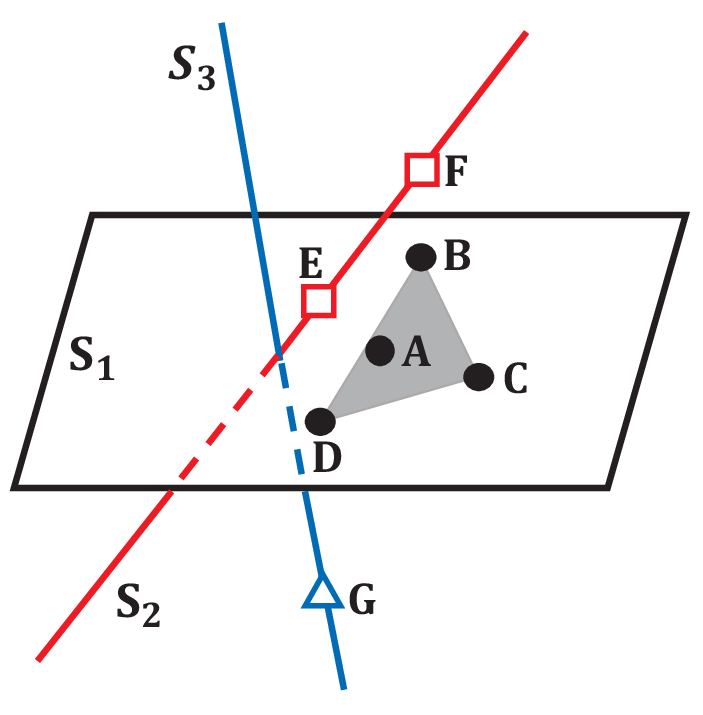}}
\caption{A key observation of the geometric analysis of three
different similarity graph construction strategies. There are
three subspaces $S_1$, $S_2$, and $S_3$ lie in $\mathbb{R}^3$,
where $dim(S_1)=2$, $dim(S_2)=1$, $dim(S_3)=1$. Points $A$, $B$,
$C$, $D$ are draw from $S_1$, point $E$, $F$ from $S_2$, and point
$G$ from $S_3$. Fig.~\ref{pairwise} shows that the most similar
point to $A$ is $E$ in terms of Euclidean distance (a kind of
PDS), but $E$ is not in the same cluster of $A$; Fig.~\ref{linear}
shows that the most similar points to $A$ are $F$ and $G$ in terms
of linear representation based similarity (i.e., LRS), because
point $A$ lies on the line spanned by $F$ and $G$;
Fig.~\ref{locallylinear} shows that our method will select $B$,
$C$ and $D$ as most similar points to $A$. Points $B$, $C$ and $D$
not only can represent $A$ with minimal residual, but are close to
$A$. They will be divided into the same cluster.} \label{fig1}
\end{figure}

In the the above problem, the first term makes $\mathbf{x}_i$ prefer to choose the near by points to represent itself; and the second term makes it produce minimal reconstruction error. Fig.~\ref{locallylinear} is a toy example showing the effectiveness of our approach.

By solving the problem (\ref{equ:optimal}), it gives that
\begin{equation}
\label{equ:solution2}
\mathbf{c}_i=\frac{\mathbf{M}_i^{-1}\mathbf{1}}{\mathbf{1}^T\mathbf{M}_i^{-1}\mathbf{1}}.
\end{equation}
where $\mathbf{M}_i=\lambda\mathbf{S}_i^T\mathbf{S}_i+(1-\lambda)(\mathbf{x}_i \mathbf{1}^T-\mathbf{D}_i)^T(\mathbf{x}_i \mathbf{1}^T-\mathbf{D}_i)$.

Note that the above solution is not sparse. It contains many trivial coefficients. This will increase the time cost of spectral clustering. By following~\cite{Peng2012}, we get a sparse similarity graph by keeping $k$ largest entries in $\mathbf{c}_i$ and setting the rests to zeros.

Once the similarity graph is built, we could apply the graph to image clustering problem under the framework of spectral clustering~\cite{Cheng2010,Shi2000,Ng2002}. Algorithm 1 summarizes the whole procedure of our algorithm.

\begin{algorithm}[!ht]
\label{algorithm1}
    \caption{Learning locally linear representation for spectral clustering}

    \begin{algorithmic}[1]
    \Require
    A given data set $\mathbf{X}\in \mathbb{R}^{m \times n}$, balance parameter $\lambda\in [0,1)$ and thresholding parameter ($k$) .

    \State  For each point $\mathbf{x}_i\in \mathbb{R}^{m}$ ($i=1, 2, \dots, n$), calculate its representation coefficients $\mathbf{c}_i\in \mathbb{R}^{n}$ by solving
    \begin{equation*}
        \begin{split}
        \min_{\mathbf{c}_i} \hspace{1mm} {\lambda \|\mathbf{S}_i\mathbf{c}_i\|_2^2+(1-\lambda)\|\mathbf{x}_i-\mathbf{D}_i\mathbf{c}_i\|_2^2}\hspace{3mm}
        \mathrm{s.t.} \hspace{1mm} \mathbf{1}^T\mathbf{c}_i=1,
        \end{split}
    \end{equation*}

    \State Remove the trivial coefficients from $\mathbf{c}_i$ by performing hard thresholding operator, i.e., keeping $k$ largest entries in $\mathbf{c}_i$ and zeroing all other elements.

    \State Construct an undirected similarity graph via $\mathbf{W}=|\mathbf{C}|+|\mathbf{C}^T|$.

    \State Perform spectral clustering~\cite{Ng2002} over $\mathbf{W}$ to get the clustering membership.

    \Ensure
    The clustering labels of the input data points.
    \end{algorithmic}
\end{algorithm}

\section{Baselines and Evaluation Metrics}

We ran the experiments over two widely-used facial image data sets, i.e., Extended Yale database B~\cite{Georgh2001} and AR database~\cite{Martinez1998}. Extended Yale database B contains 2014 near frontal face images of 38 individuals. AR database contains 1400 face images without disguises distributed over 100 individuals (14 images for each subject). We downsized the images of Extended Yale database B from $192\times 168$ to $48\times 42$ and the AR images from $165\times 120$ to $55\times 40$. Moreover, as did in~\cite{Elham2013,Liu2013}, Principal Component Analysis (PCA) is used as a pre-processing step by retaining $98\%$ energy of the cropped images.

We compared LLR with several state-of-the-art algorithms, i.e., LRR~\cite{Liu2013}, SSC~\cite{Elham2013}, LLE-graph based Clustering (LLEC)~\cite{Roweis2000}, and standard Spectral Clustering (SC)~\cite{Ng2002}. Moreover, we also tested the performance of k-means clustering as a baseline.

Two popular metrics, Accuracy (AC) and Normalized Mutual Information (NMI), are used to measure the clustering performance of these algorithms. The method works better, the value of AC or NMI is higher. In addition, the time cost for building similarity graph ($t_1$) and the whole time cost for clustering ($t_2$) are recorded to evaluate the efficiency.

In each test, we tuned the parameters of all the methods to obtain their best AC. In details, LLR needs two user-specified parameters, balance parameter $\lambda$ and thresholding parameter $k$. We set $\lambda \in \{0.001,0.01,0.1\}$ and $k \in \{3,4,5,6\}$. Moreover, considering the computation efficiency, we only use $300$-nearest data points as dictionary $\mathbf{D}_i$ for each $\mathbf{x}_i$ in terms of Euclidean distance. For the other compared methods, we set the parameters by following~\cite{Liu2013,Elham2013,Roweis2000,Ng2002}.

\begin{table}[t]
\label{tab1}
\centering
\caption{performance comparisons in different methods over Extended Yale database B. $t_1$ denotes the CPU elapse time (second) for building similarity graph and $t_2$ is the whole time cost.}
{\begin{tabular}{ccccccc}
\hline
Metric & {LLR}&{LRR~\cite{Liu2013}} &{SSC~\cite{Elham2013}} & {LLEC~\cite{Roweis2000}} &{SC~\cite{Ng2002}} &k-means\\
\hline
AC &\textbf{0.883}& 0.713& 0.613& 0.461& 0.426 &0.098\\

NMI & \textbf{0.922} & 0.772 & 0.684 & 0.540 & 0.539 & 0.115\\

$t_1$ & 14.628 & 38.095 & 159.665 &0.678& 0.264 &-\\

$t_2$ & 102.256 & 90.8268 & 231.235 &74.309& 64.606 &\textbf{4.543}\\\cline{1-7}
\end{tabular}}{}
\end{table}

\begin{table}[t]
\label{tab2}
\centering
\caption{performance comparisons in different methods over AR database.}
{\begin{tabular}{ccccccc}
\hline
Metric & {LLR}&{LRR~\cite{Liu2013}} &{SSC~\cite{Elham2013}} & {LLEC~\cite{Roweis2000}} &{SC~\cite{Ng2002}} &k-means\\
\hline
AC &\textbf{0.837} &0.771 & 0.767& 0.396& 0.361 &0.311\\

NMI & \textbf{0.929} & 0.910 & 0.886 &0.682& 0.652 &0.611\\

$t_1$ & 8.696 & 30.495 & 164.327 &0.318& 0.147 &-\\

$t_2$ & 111.618 & 128.343 & 286.978 &107.779& 113.918 &\textbf{4.460}\\\cline{1-7}
\end{tabular}}{}
\end{table}

We report the clustering results of the evaluate algorithms in Table $1$ and Table $2$, from which we have the following observations:
\begin{itemize}
  \item LLR outperforms the other methods in AC and NMI by a considerable performance margin. LLR is $6.6\%$ and $1.9\%$ higher than the second best method (LRR) over AR in AC and NMI, respectively. The corresponding values are $17.0\%$ and $15.0\%$ over Extended Yale Database B.
  \item LRR and SSC are two recently-proposed algorithms, which are superior to LLEC and SC. Note that, only SC is a pairwise distance-based spectral clustering method.
  \item LLR finds an elegant balance between time cost and clustering quality, which is not the fastest algorithm but achieves the best clustering quality.
  \item k-means is the fastest algorithm, but performs the worst in AC and NMI.
\end{itemize}

\section{Conclusion}

Linear representation and pairwise distance are two popular methods to construct a similarity graph for spectral clustering. But both of them encountered some problems in practical applications. Pairwise distance-based method is sensitive to noise and outliers, while linear representation-based method might fail when the data came from a union of dependent subspaces. In this letter, we proposed a new algorithm that represents the objective point $\mathbf{x}$ using some data points that not only can reconstruct $\mathbf{x}$ better but also are close to $\mathbf{x}$ in terms of pairwise distance. The incorporation of pairwise distance and linear representation largely improve the discrimination of data model, which is beneficial to clustering problem. Extensive experiments have verified the effectiveness and efficiency of our approach.

\section*{Acknowledge}
{This work was supported by the National Basic Research Program of China (973 Program) under grant 2011CB302201, the Program for New Century Excellent Talents in University of China under Grant NCET-12-0384, and the National Natural Science Foundation of China under Grant 61172180.}


\begin{thebibliography}{}

\bibitem{Cheng2010}
Cheng, B., Yang, J., Yan, S., Fu, Y. and Huang, T.: `Learning with $\ell^1$-graph for image analysis', IEEE Trans. Image proc., 2010, \textbf{19}, (4), pp. 858-866


\bibitem{Belkin2003}
Belkin, M., and Niyogi, P.: `Laplacian eigenmaps for dimensionality reduction and data representation', \textit{Neural computation}, 2003, \textbf{15}, (6), pp. 1373-1396


\bibitem{Elham2013}
Elhamifar, E., and Vidal, R.: `Sparse subspace clustering: Algorithm, theory, and applications', IEEE Trans. Pattern Anal. Mach. Intell., 2013, \textbf{35}, (11), pp. 2765-2781


\bibitem{Roweis2000}
S. Roweis, L. Saul: `Nonlinear dimensionality reduction by locally linear embedding', \textit{Science}, 2000, \textbf{290}, (5500), pp. 2323-2326


\bibitem{Liu2013}
Liu, G., Lin, Z., Yan, S., Sun, J., Yu, Y., and Ma, Y.: `Robust recovery of subspace structures by low-rank representation', IEEE Transactions on Pattern Anal. and Mach. Intell., 2013, \textbf{35}, (1), pp. 171-184




\bibitem{Peng2012}
Peng, X., Zhang, L., Zhang, Y.: `Constructing L2-Graph For Subspace Learning and Segmentation', preprint  arXiv:1209.0841v4, 2012

\bibitem{Shi2000}
Shi, J. and Malik, J.: `Normalized cuts and image segmentation', IEEE Trans. Pattern Anal. Mach. Intell., 2000, \textbf{22}, (8), pp. 888-905

\bibitem{Ng2002}
Ng, A. Y., Jordan, M. I., and Weiss, Y.: `On spectral clustering: Analysis and an algorithm', Advances in Neural Information Processing Systems (NIPS), 2002, pp. 849-856

\bibitem{Georgh2001}
Georghiades, A., and Belhumeur,P., and Kriegman, D.: `From few to many: Illumination cone models for face recognition under variable lighting and pose', IEEE Trans. Pattern Anal. Mach. Intell., 2001, \textbf{23}, (6), pp. 643-660

\bibitem{Martinez1998}
Martinez, A., and Benavente, R.: `The AR face database', CVC Tech. Report No. 24, 1998





\end{thebibliography}
\end{document}